\documentclass[letterpaper, 10 pt, conference]{ieeeconf}  

\usepackage{color,graphicx}
\usepackage{subfig}
\usepackage{csquotes}
\usepackage{framed}
\usepackage{amsmath}
\usepackage[font={small}]{caption}
\usepackage{hhline}

\makeatletter
\let\NAT@parse\undefined
\makeatother
\usepackage{hyperref}

\usepackage[belowskip=-6pt,aboveskip=2pt]{caption}

\IEEEoverridecommandlockouts                              

\usepackage{graphics} 

\title{\LARGE \bf
Iterative Hough Forest with Histogram of Control Points for 6 DoF Object Registration from Depth Images
}

\author{Caner Sahin, Rigas Kouskouridas and Tae-Kyun Kim
\thanks{All authors are with the Imperial Computer Vision and Learning Lab (ICVL), at the Department of Electrical and Electronic Engineering, Imperial College London, UK, {\tt\small \{c.sahin14, r.kouskouridas, tk.kim\}@imperial.ac.uk}}%
}

\begin{document}

\maketitle
\thispagestyle{empty}
\pagestyle{empty}

\begin{abstract}
State-of-the-art techniques proposed for 6D object pose recovery depend on occlusion-free point clouds to accurately register objects in 3D space. To reduce this dependency, we introduce a novel architecture called \textit{Iterative Hough Forest with Histogram of Control Points} that is capable of estimating occluded and cluttered objects' 6D pose given a candidate 2D bounding box. Our \textit{Iterative Hough Forest} is learnt using patches extracted only from the positive samples. These patches are represented with \textit{Histogram of Control Points (HoCP)}, a \enquote{scale-variant} implicit volumetric description, which we derive from recently introduced Implicit B-Splines (IBS). The rich discriminative information provided by this scale-variance is leveraged during inference, where the initial pose estimation of the object is iteratively refined based on more discriminative control points by using our \textit{Iterative Hough Forest}. We conduct experiments on several test objects of a publicly available dataset to test our architecture and to compare with the state-of-the-art.
\end{abstract}
\section{Introduction}
Object registration is an important task in computer vision that determines the translation and the rotation of an object with respect to a reference coordinate frame. By utilizing such a task, one can propose promising solutions for various problems related to scene understanding, augmented reality, control and navigation of robotics, \textit{etc}. Recent developments on visual depth sensors and their increasing ubiquity have allowed researchers to make use of the information acquired from these devices to facilitate the registration.\\
\indent When the target point cloud is cleanly segmented, Iterative Closest Point (ICP) algorithm \cite{1}, point-to-model based methods \cite{2, 3} and point-to-point techniques \cite{4, 5} demonstrate good results. However, the performances of these approaches are severely degraded by the challenges such as heavy occlusion and clutter, and similar looking distractors. In order to address these challenges, several learning based methods formulate occlusion aware features \cite{6}, derive patch-based (local) descriptors \cite{7} or encode the contextual information of the objects with simple depth pixels \cite{8} and integrate into random forests. Particularly, iterative random forest algorithms such as Latent-Class Hough forest (LCHF) \cite{7} and iterative Multi-Output Random forest (iMORF) \cite{9} obtain the state-of-the-art accuracy on pose estimation. On the other hand, these methods rely on scale-invariant features and the exploitation of the rich discriminative information that is inherently embedded into the scale-variability is one important point overlooked.\\
\begin{figure}[!t]
\captionsetup[subfigure]{labelformat=empty}
\centering
\includegraphics[height=2in]{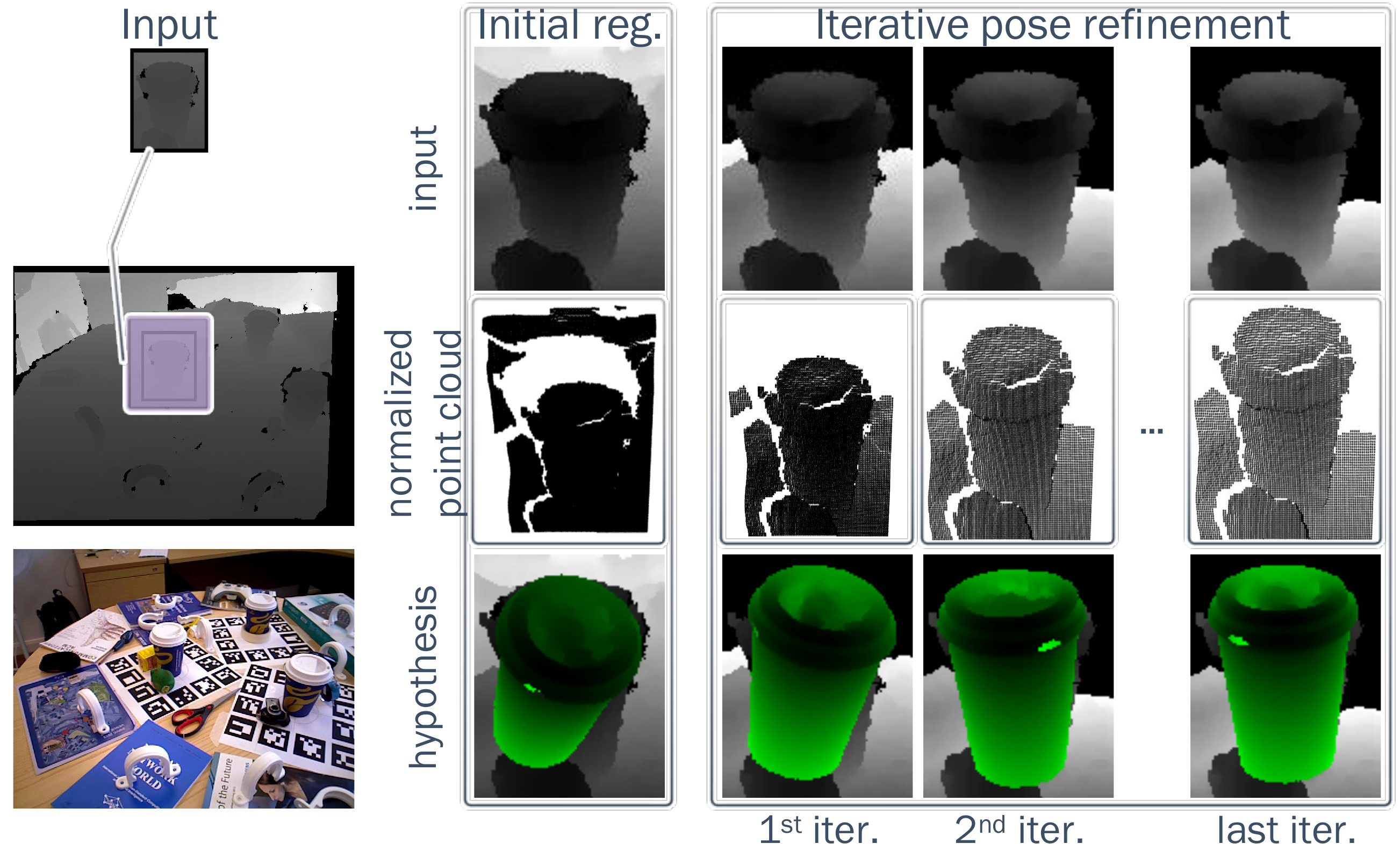}
\caption{Sample result of our architecture: \textit{initial registration} roughly aligns the test object and \textit{iterative pose refinement} further refines this alignment (The RGB image is for better visualization).}
\vspace{-2em}
\label{fig1}
\end{figure}
\indent Unlike the aforementioned learning-based methods, Novatnack \textit{et al.} \cite{10, 11} utilize the detailed information of the scale variation in order to register the range images in a coarse-to-fine fashion. They extract and match conventional salient 3D key points. However, real depth sensors have several imperfections such as missing depth values, noisy measurements, foreground fattening, \textit{etc}. Salient feature points tend to be located on these deficient parts of the depth images, and hence, they are rather unstable \cite{12}. In such a scenario, 3D reconstruction methods that provide more reliable shape information can be utilized \cite{6}. Implicit B-Splines (IBS) \cite{3, 13} are yet other techniques that can provide shape descriptors through their zero-sets and reconstruct surfaces. These techniques are based on the locally controlled functions that are combined via their control points and this local control allows patch-based object representation.\\
\indent Our architecture is originated from these observations. We integrate the coarse-to-fine registration approach presented in \cite{10} into the random forests \cite{7, 9} using the Histogram of the Control Points (HoCP) that we adapt from recently introduced IBSs \cite{13}. We train our forest only from positive samples and learn the detailed information of the scale-variability during training. We normalize every training point cloud into a unit cube and then generate a set of scale-space images, each of which is separated by a constant factor. The patches extracted from the images in this set are represented with the scale-variant HoCP features. During inference, the patches centred on the pixels that belong to the background and foreground clutters are removed iteratively using the most confident hypotheses and the test image is updated. Since this removal process decreases the standard deviation of the test point cloud, subsequent normalization applied to the updated test image increases the relative scale of the object (foreground pixels) in the unit cube. More discriminative descriptors (control points) are computed at higher scales and this ensures further refinement of the object pose. Note that we employ a compositional approach, that is, we concurrently detect the object in the target region and estimate its pose by aligning the patches in order to increase robustness across clutter. Figure \ref{fig1} depicts a sample result of our architecture. To summarize, our main contributions are as follows:
\begin{itemize}
\item To the best of our knowledge this is the first time we adapt an implicit object representation, Implicit B-Spline, into a \enquote{scale-variant} patch descriptor and associate with the random forests.
\item We introduce a novel iterative algorithm for the Hough forests: it finds out an initial hypothesis and improves its confidence iteratively by extracting more discriminative \enquote{scale-variant} descriptors due to the elimination of the background/foreground clutter.
\end{itemize}
\section{Related Work}
\label{Related Work}
A large number of studies have been proposed for the object registration, ranging from the point-wise correspondence based methods to the learning based approaches. Iterative Closest Point (ICP) algorithm, originally presented in \cite{1}, requires a good initialization in order not to be trapped in a local minimum during fine tuning. This requirement is reduced in \cite{14} providing globally optimal registration by the integration of a global branch-and-bound (BnB) optimization into the local ICP. The point-wise correspondence problem is converted into a point-to-model registration in \cite{2, 3}. The object model is represented with implicit polynomials (IP) and the distance between the test point set and this model is minimized via the Levenberg-Marquardt algorithm (LMA). The study \cite{15} that utilizes 3D IPs for 6 DoF pose estimation on ultrasound images is further extended in \cite{16} by a coarse-to-fine IP-driven registration strategy. The point-to-point techniques build point-pair features for sparse representations of the test and the model point sets \cite{17}. Rusu \textit{et al.} align two noisy point clouds of real scenes by finding correct point-to-point correspondences between the Point Feature Histograms (PFH) and feed this alignment to an ICP algorithm for fine tuning \cite{18}. The votes of the matching features are accumulated in \cite{19} to hypothesise the poses of the cluttered and partially occluded objects. Choi \textit{et al.} \cite{20} propose point-pair features for both RGB and depth channels and they are conducted in a voting scheme to hypothesise the rotation and translation parameters of the objects in the cluttered scenes. Despite achieving good registration results, these techniques underperform when the scenes are under heavy occlusion and clutter, and the target objects' geometry are indistinguishable from background clutter.\\
\indent Learning-based methods have good generalization across severe occlusion and clutter \cite{24}. The state-of-the-art accuracy on registration is acquired by the iterative random forest algorithms, particularly \cite{7} and \cite{9}, which form a basis for our \textit{Iterative Hough Forest} architecture. Tejani's patch-based strategy \cite{7} refines the initially hypothesised object pose by iteratively updating the object class distributions in the leaf nodes during testing. Iterative Multi Output Random forest (iMORF) \cite{9} jointly predicts the head pose, the facial expression and the landmark positions. The relations between these tasks are modelled so that their performances are iteratively improved with the extraction of more informative features. Whilst these approaches rely on the scale-invariant features to improve the confidence of a pose hypothesis, inspiring by \cite{10}, we design scale-variant features getting more discriminative with the increase in the scale. Novatnack et al. \cite{10, 11} introduce a framework that registers the range images in a coarse-to-fine fashion by utilizing the detailed information provided by the scale variation. The shape descriptors with the coarsest scale are matched initially and a rough alignment is achieved since fewer features are extracted in coarser scales. The descriptor matching at higher scales results improved predictions of the pose. 
\section{Our Registration Approach}
\label{approach}
In this section we detail our registration approach by firstly describing the computation procedure of the HoCP features. We then present how to encode the discriminative information of these scale-variant features into the forest. Finally, we demonstrate how to exploit the learnt shape information in a coarse-to-fine fashion to refine the pose hypotheses.
\subsection{Histogram of Control Points (HoCP)}
\begin{figure*}[!t]
\captionsetup[subfigure]{labelformat=empty}
\centering
\includegraphics[height=1.55in]{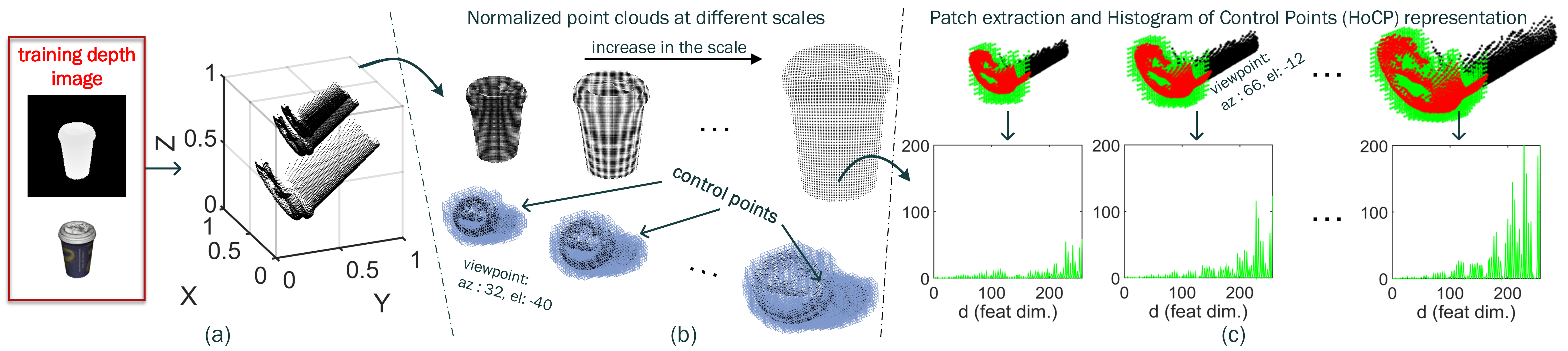}
\caption{Computation procedure of the HoCP features as a scale-variant patch representation: (a) initial normalization ($s_0 = 1$) of the training depth image is the outmost point cloud and the inner ones are sampled by different $s_i$ values. (b) global representations of the scale-space images. (c) HoCP representation of the sample part (red).}
\label{fig2}
\vspace{-1em}
\end{figure*}

We demonstrate the computation procedure of the HoCP features over a positive depth image selected from the training dataset. It is initially normalized into a unit cube and then new point clouds at different scales are sampled as follows:
\begin{equation}
  \{ {\mathbf{X}_N \}}_i= \frac{\mathbf{X}_{n \times 3} - \bar{\mathbf{X}}_{n \times 3}}{s_i * \mathbf{\alpha}} + 0.5, \quad i = 0, 1, 2, ..., m
  \label{eq1}
\end{equation}
with
\begin{equation}
\small
\mathbf{\alpha} = \max \left  \{
				  \begin{tabular}{ccc}
				  max($X$)-min($X$) \\
				  max($Y$)-min($Y$) \\
				  max($Z$)-min($Z$) 
				  \end{tabular}
					  \right \}, h_i = \max(Z_{N_i})-\min(Z_{N_i})
					  \label{eq2}
\end{equation}
where $\textbf{X} = [X, Y, Z]$ is the world coordinate vector of the original foreground point cloud, $\bar{\mathbf{X}}$ is the mean of $\textbf{X}$, $\textbf{X}_N = [X_N,  Y_N,  Z_N]$ is the normalized foreground pixels, $m$ is the number of the scales, $\alpha$ is the scale factor and $h$ is the scale. The constant $s_i$ takes real numbers to generate the point clouds at different scales, starting from $s_0 = 1$ that corresponds to the initial normalization. A training image and its samples at different scales are shown in Fig. \ref{fig2} (a).\\
\indent Once we generate a set of scale-space images (Fig. \ref{fig2} (a)), we represent these point clouds with the control point descriptors first globally. The descriptor computation procedure is the same as presented in \cite{13}. The unit cube is split into an $ N \times N \times N $ voxel grid where $N$ is the IBS resolution. Each descriptor $\Gamma$ is defined with an index-weight pair: the index number indicates the vertex of this grid at which the related control point is located. The weight informs the descriptor significance about the control of the geometry to be represented. The scale-space images in Fig. \ref{fig2} (a) are globally represented in Fig. \ref{fig2} (b). We next partition the global representation at each scale into patches. We express the patch size $g$ in image pixels and it is a constant that depicts the ratio between the sizes of the extracted patch and the bounding box of the global point cloud. A window with the specified patch size is traversed in the unit cube of each scale-space image and the patches are extracted around non-zero pixels. Each patch has its own implicit volumetric representation, formed by the closest control points to the patch center, the ones lying inside the window along depth direction. The patches sampled at different scales in Fig. \ref{fig2} (c) represent the same shape. However, their volumetric descriptions (green) are getting more discriminative as the scale increases, since the greater number of descriptors are computed at higher scales. We encode this discriminative information into histograms in spherical coordinates. Each of the patch centres is coincided with the center of a sphere. The control points of the patch are described by the log of the radius $t_r$, the cosine of the inclination $t_{\theta}$ and the azimuth $t_{\phi}$. Then, the sphere is divided into the bins and the relation between the bin numbers $h_r, h_{\theta}, h_{\phi}$ and the histogram coordinates $t_r, t_{\theta}, t_{\phi}$ is given as follows \cite{21}:
\vspace{-0.5em}
\begin{equation}
\begin{split}
       t_r &= \frac{h_r}{log(\frac{r_{max}}{r_{min}})} log(\frac{r}{r_{min}}) \\
t_{\theta} &= h_{\theta} \frac{z}{r} \\
  t_{\phi} &= \frac{h_{\phi}}{2 \pi} tan^{-1}(\frac{y}{x})  \\
\end{split}
\end{equation}
where $r_{min}$ and $r_{max}$ are the radii of the nested spheres with the minimum and the maximum volumes, $x, y, z$ are the Cartesian coordinates of each descriptor with radius $r$. $r_{max}$ equals to the distance between the patch center and the farthest descriptor of the related patch. The numbers of the control points in each bin are counted and stored in a $d = h_r*h_{\theta}*h_{\phi}$ dimensional feature vector $\mathbf{f}$. The volumetric descriptions in Fig. \ref{fig2} (c) are shown with their related histograms. Thus, the sample shape (patch) is represented with the scale-variant HoCP features.
\subsection{The Combination of HoCP and Iterative Hough Forest}
The proposed Iterative Hough Forest is the combination of randomized binary decision trees. It is trained only on foreground synthetically rendered depth images of the object of interest. We generate a set of scale-space images from each training point cloud and sample a set of annotated patches $\{ \cup_{i=1}^p P_i \}$ as follows \cite{7}:
\begin{equation}
\mathcal{P} = \{ \cup_{i=1}^p P_i \} = \{ \cup_{i=1}^p (\mathbf{c}_i, \Delta \mathbf{x}_i, \mathbf{\theta}_i, \mathbf{f}_i, D_i)\}
\end{equation}
where $\mathbf{c}_i = (c_{x_i}, c_{y_i})$ is the patch centre in pixels, $ \Delta \mathbf{x}_i = (\Delta x_i, \Delta y_i, \Delta z_i)$ is the 3D offset between the centres of the patch and the object, $\theta_i = (\theta_{r_i}, \theta_{p_i}, \theta_{y_i}) $ is the rotation parameters of the point cloud from which the patch $P_i$ is extracted and $D_i$ is the depth map of the patch.\\
\indent Each tree is constructed by using a subset $\mathcal{S}$ of the annotated training patches $\mathcal{S} \subset \mathcal{P}$. We randomly select a template patch $T$ from $\mathcal{S}$ and assign it to the root node. We measure the similarity between $T$ and each patch $S_i$ in $\mathcal{S}$ as follows:
\begin{itemize}
\item \textbf{Depth check:} The depth values of the descriptors $S_{i_\Gamma}$ and $T_{\Gamma}$ that represent the patches $S_i$ and $T$ are checked, and the spatially inconsistent ones in $S_{i_\Gamma}$ are removed as in \cite{7}, generating $\Omega$ that includes the spatially consistent descriptors of the patch $S_i$.
\item \textbf{Similarity measure:} Using $\Omega$, the feature vector $\mathbf{f}_{\Omega}$ is generated and the $\mathcal{L}_2$ norm between this vector and $\mathbf{f}_T$ is measured:
\begin{equation}
\mathcal{F}(S_i, T) = {\parallel \mathbf{f}_{\Omega} - \mathbf{f}_T \parallel}_2
\end{equation}
\item \textbf{Similarity score comparison:} Each patch is passed either to the left or the right child nodes according to the split function that compares the score of the similarity measure $\mathcal{F}(S_i, T)$ and a randomly chosen threshold $\tau$.
\end{itemize}
\indent The main reason why we apply a depth check to the patches is to remove the structural perturbations, due to occlusion, clutter \cite{7}. These perturbations most likely occur on the patches extracted along depth discontinuities such as the contours of the objects of interest. They cause to diverge a test patch (occluded/cluttered) from its positive correspondence by changing its representation, $r_{max}$ of the sphere, and the histogram coordinates consequently.\\
\indent A group of candidate split functions are produced at each node by using a set of randomly assigned patches $\{ T_i \}$ and thresholds $\{ \tau_i \}$. The one that best optimize the offset and pose regression entropy \cite{22} is selected as the split function. Each tree is grown by repeating this process recursively until the forest termination criteria are satisfied. When the termination conditions are met, the leaf nodes are formed and they store votes for both the object center $\Delta \mathbf{x} = (\Delta x, \Delta y, \Delta z)$ and the object rotation $\theta = (\theta_r, \theta_p, \theta_y) $.
\subsection{Initial Registration and Iterative Pose Refinement}
\begin{figure}[!t]
\captionsetup[subfigure]{labelformat=empty}
\centering
\includegraphics[height=3.9in]{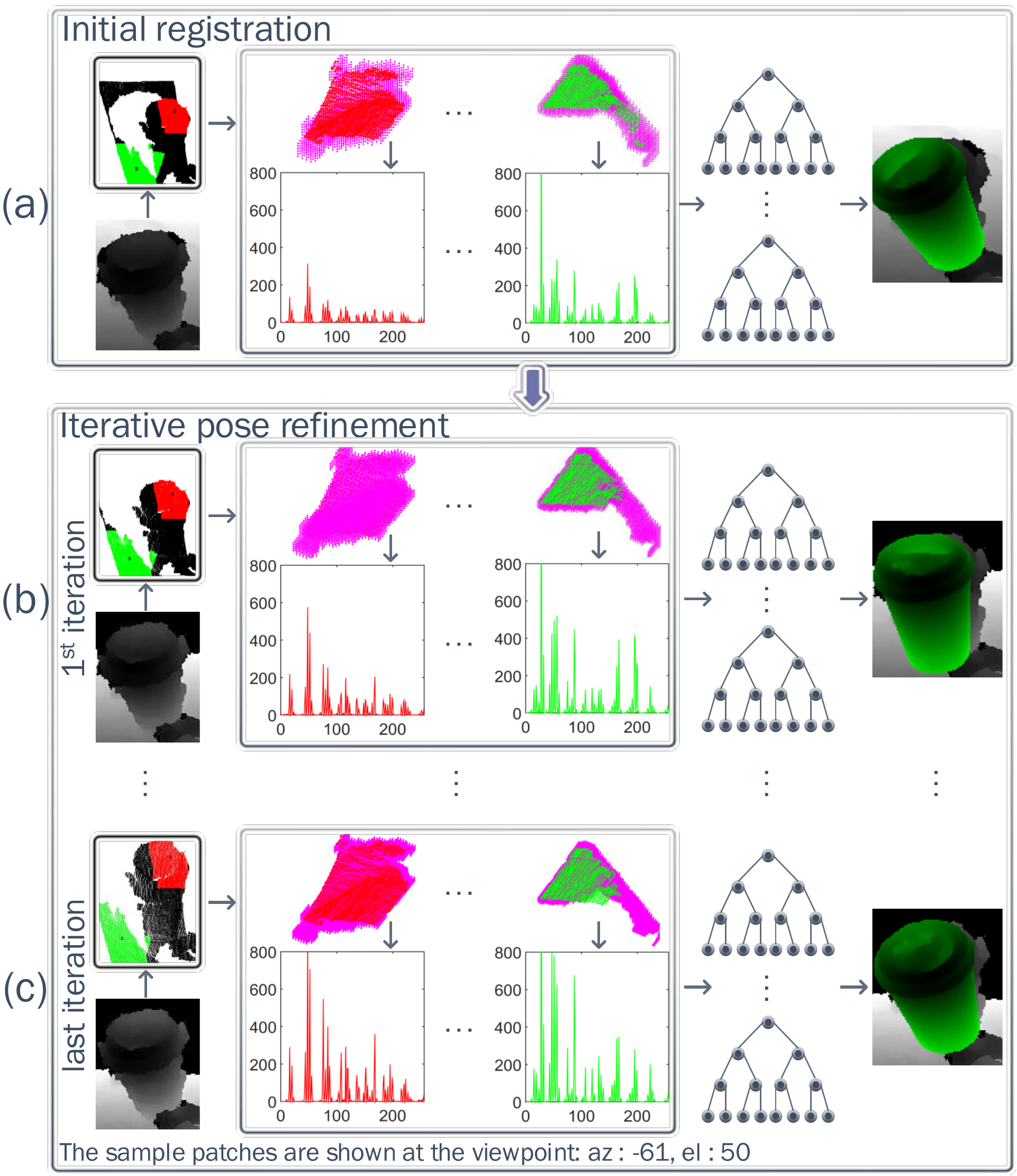}
\caption{Initially estimated object pose is iteratively refined based on more discriminative control points that are extracted due to the elimination of background/foreground clutter.}
\label{fig3}
\end{figure}

The proposed architecture registers the object of interest in two steps: the \textit{initial registration} and the \textit{iterative pose refinement}. The \textit{initial registration} roughly aligns the test object and this alignment is further improved by the \textit{iterative pose refinement}.\\
\indent Consider an object that was detected by a coarse bounding box, $I_b$, as shown in the leftmost image of Fig. \ref{fig3} (a). At an iteration instant $k$, the following quantities are defined:
\begin{itemize}
\item $\Delta \mathbf{x}^{0:k} = \{ \Delta \mathbf{x}^0, \Delta \mathbf{x}^1, ..., \Delta \mathbf{x}^k \} = \{ \Delta \mathbf{x}^0, \Delta \mathbf{x}^{1:k} \}$: the history of the object position.
\item $\theta^{0:k} = \{ \theta^0, \theta^1, ..., \theta^k \} = \{ \theta^0, \theta^{1:k} \}$: the history of the object rotation.
\item $V^{1:k} = \{ v^1, v^2, ..., v^k \}$ : the history of the inputs (noise removals) applied to the test image.
\item $m^{0:k} = \{ m^0, m^1, ..., m^k \} = \{ m^0, m^{1:k} \}$: the history of the set of the feature vectors where $m^k = \{ \cup_{i=1}^n \mathbf{f}_i \}$.
\item $h^k$: the object scale (the scale of the foreground pixels) in the unit cube at iteration $k$ (see Eq. \ref{eq2}).
\end{itemize}
We formulate the \textit{initial registration} as follows:
\begin{equation}
( \Delta \mathbf{x}^0, \theta^0 ) = \arg \max_{\Delta \mathbf{x}^0, \theta^0}  p(\Delta \mathbf{x}^0, \theta^0 \vert I_b, m^0, h^0).
\label{eq8}
\end{equation}
We find the best parameters that maximize the joint posterior density of the initial object position $\Delta \mathbf{x}^0$ and the initial object rotation $\theta^0$. This initial registration process is illustrated in Fig. \ref{fig3} (a). The test image is firstly normalized into a unit cube. Unlike training, this is a \enquote{single} scale normalization that corresponds to $s_0 = 1$ (see Eq. \ref{eq1}). The patches extracted from the globally represented point cloud are described with the HoCP features and passed down all the trees. We determine the effect that all patches have on the object pose by accumulating the votes stored in the leaf nodes as in \cite{7} and approximate the initial registration given in Eq. \ref{eq8}. Once the initial hypothesis $ \mathbf{x}^0 = (\Delta \mathbf{x}^0, \theta^0)$ is obtained, the pixels that belong to the background/foreground clutter $\{ \cup_{j = 0}^f p_j \}$ are removed from $I_b$ according to the following criterion:
\begin{equation}
    v^k=
    \begin{cases}
      I_b(p_j) = \mathcal{D}_{I_b}(p_j), & \gamma \psi_1 < \mathcal{D}_{I_b} (p_j^k) < \beta \psi_2 \\
      I_b(p_j) = 0, & otherwise 
    \end{cases}
    \label{eq10}
  \end{equation}
  with
  \begin{equation}
  \gamma = \min (\mathcal{D}_H^k), \quad \beta = \max (\mathcal{D}_H^k)
  \end{equation}
where $\mathcal{D}_H^k$ and $\mathcal{D}_{I_b}$ are the depth maps of the hypothesis $H$ at iteration $k$, and of the $I_b$, $\psi_1$ and $\psi_2$ are the scaling coefficients. The efficacy of $v^k$ is illustrated in Fig. \ref{fig3}. In the rightmost image of Fig. \ref{fig3} (a), the test image and the initial hypothesis are overlaid. This hypothesis is exploited and the test image is updated by $v^1$ as in Eq. \ref{eq10}. This updated image is shown in Fig. \ref{fig3} (b) and assigned as input for the $1^{st}$ iteration. It is normalized and represented globally. Note how the object \enquote{scale} ($h^1$) in the unit cube is relatively increased and more discriminative descriptors $m^1$ are extracted (compare with the initial registration). This is mainly because of that the standard deviation of the input image is decreased since we removed foreground/background clutter. The resultant hypothesis of the $1^{st}$ iteration is shown on the right. The extraction of more discriminative descriptors and the noise removal process result more accurate and confident hypothesis. This pose refinement process is iteratively performed until the maximum iteration is reached (see Fig. \ref{fig3} (c)):
\begin{equation}
( \Delta \mathbf{x}^k, \theta^k ) = \arg \max_{\Delta \mathbf{x}^k, \theta^k}  p(\Delta \mathbf{x}^k, \theta^k \mid m^{1:k}, V^{1:k}, \mathbf{x}^0, h^k)
\end{equation}
We approximate the registration hypothesis at each iteration by using the stored information in the leaf nodes as we do in the initial registration.
\section{Experimental Results}
\label{experiments}
We have analysed the ICVL dataset \cite{7} and have found that the \enquote{coffee cup} and the \enquote{camera} are some of the best demonstrable objects to test and compare our registration architecture with the state-of-the-art methods since they are located in highly occluded and cluttered scenes. We further process the test images of these objects to generate a new test dataset according to the following criteria:
\begin{itemize}
\item Since the HoCP features are scale-variant, the depth values of the training and the test images should be close to each other up to a certain degree. In this study, we train the forests at a single depth value, $750$ mm, and test with the images at the range of $[750 \mp 35]$ mm.
\item The test object instances located at the range of $[750 \mp 35]$ mm are assumed as detected by coarse bounding boxes (see Fig. \ref{fig1}). The image regions included in these bounding boxes are cropped and the new test dataset is generated ($276$ \enquote{coffee cup} and $360$ \enquote{camera} RGBD test images).
\end{itemize}
The maximum depth is $25$ and the number of the maximum samples at each leaf node is $15$ for each tree. Every forest is the ensemble of $3$ trees with these termination criteria. Our experiments are two folds: \textit{intraclass} and \textit{interclass}. Both experiments use the metric proposed in \cite{23} to determine whether a registration hypothesis is correct. This metric outputs a score $\omega$ that calculates the distance between the ground truth and estimated poses of the test object. The registration hypothesis that ensures the following inequality is considered as correct:
\begin{equation}
\omega \leq z_{\omega} \Phi
\end{equation}
where $\Phi$ is the diameter of the 3D model of the test object and $z_{\omega}$ is a constant that determines the coarseness of an hypothesis that is assigned as correct. We set $z_{\omega}$ to $0.08$ in the intraclass and interclass experiments.
\subsection{Intraclass Experiments}
\begin{figure}[!t]
\captionsetup[subfigure]{labelformat=empty}
\centering
\includegraphics[height=1.8in]{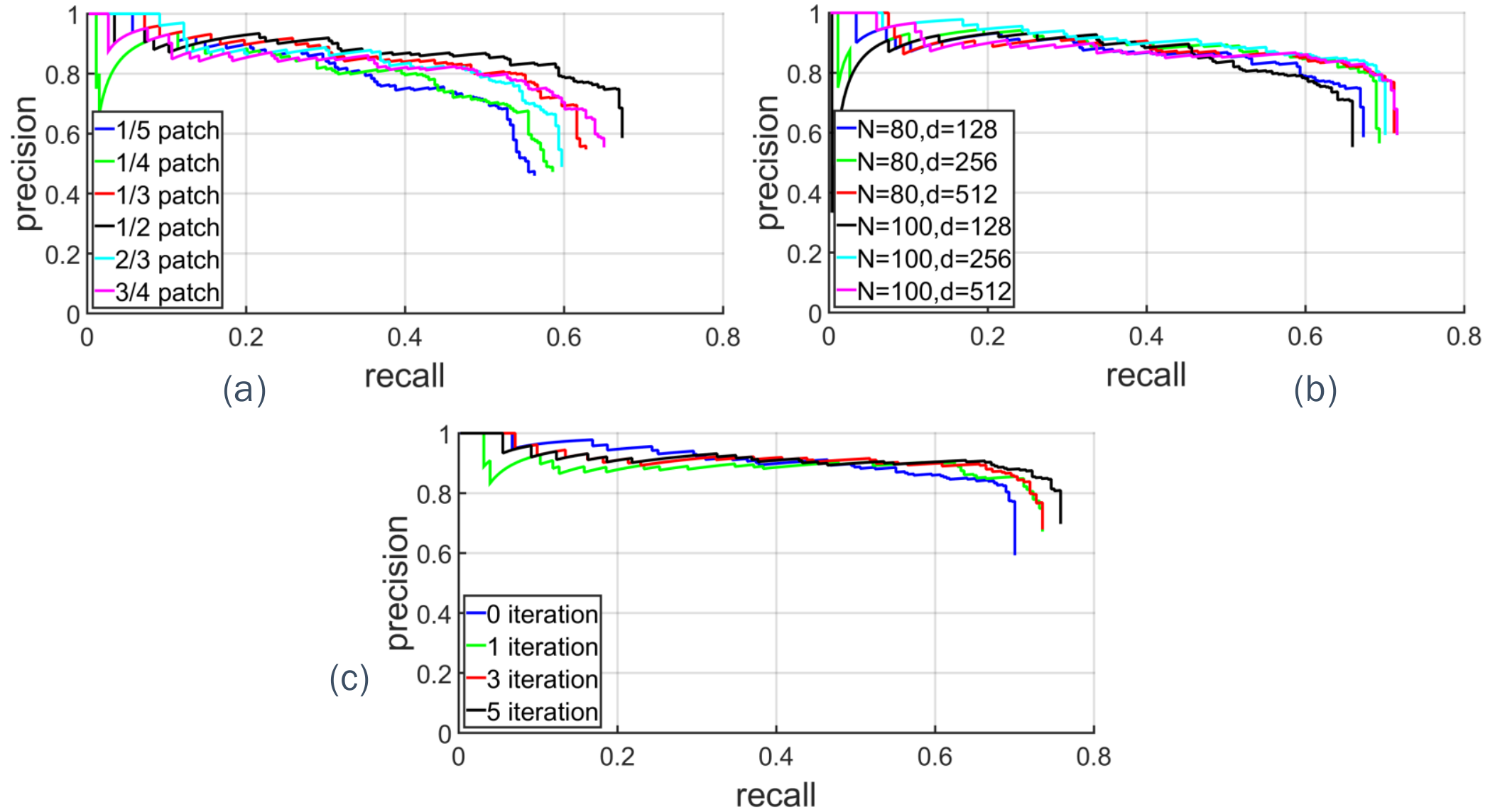}
\caption{PR curves obtained from the intraclass experiments: according to these results, we choose $\frac{1}{2}$ patch size and set $N = 100$, $d = 256$. For the corresponding F1 scores, see Table \ref{table_params}.}
\label{fig4}
\end{figure}

These experiments are performed on the \enquote{coffee cup} dataset to determine the optimal parameters of the proposed approach. The effect of the patch size $g$ is firstly examined by setting the IBS resolution $N$ to $80$, the HoCP feature dimension $d$ to $128$ in addition to the previously defined forest parameters. We test the patch sizes $g = \{ 0.20, 0.25, 0.33, 0.50, 0.66 , 0.75 \}$. The resultant Precision-Recall (PR) curves are shown in Fig. \ref{fig4} (a). When we increase the patch size until it is $0.5$ times of the bounding box, the registration performance is improved since the greater patches can encode more discriminative shapes. We continue to extend the patch size till it is $0.75$ times of the bounding box and observe that the performance is degraded since these patches tend to contain the noisy parts of the scene. According to this figure and their corresponding F1 scores (see Table \ref{table_params}), we choose $\frac{1}{2}$ as the optimal patch size.\\
\indent By using the selected patch size, we next tune the IBS resolution $N$ and the HoCP feature dimension $d$. We test the combinations of $N = \{80, 100\}$ and $d = \{128, 256, 512\}$, the ones that are the most applicable $N-d$ pairs to represent $\frac{1}{2}$ patch size. The PR curves of these combinations are depicted in Fig. \ref{fig4} (b) and the corresponding F1 scores are illustrated in Table \ref{table_params}. We take into account both the memory consumption and the accuracy, and agree on the values of $N = 100$ \& $d = 256$.
\begin{figure}[!t]
\captionsetup[subfigure]{labelformat=empty}
\centering
\includegraphics[height=1in]{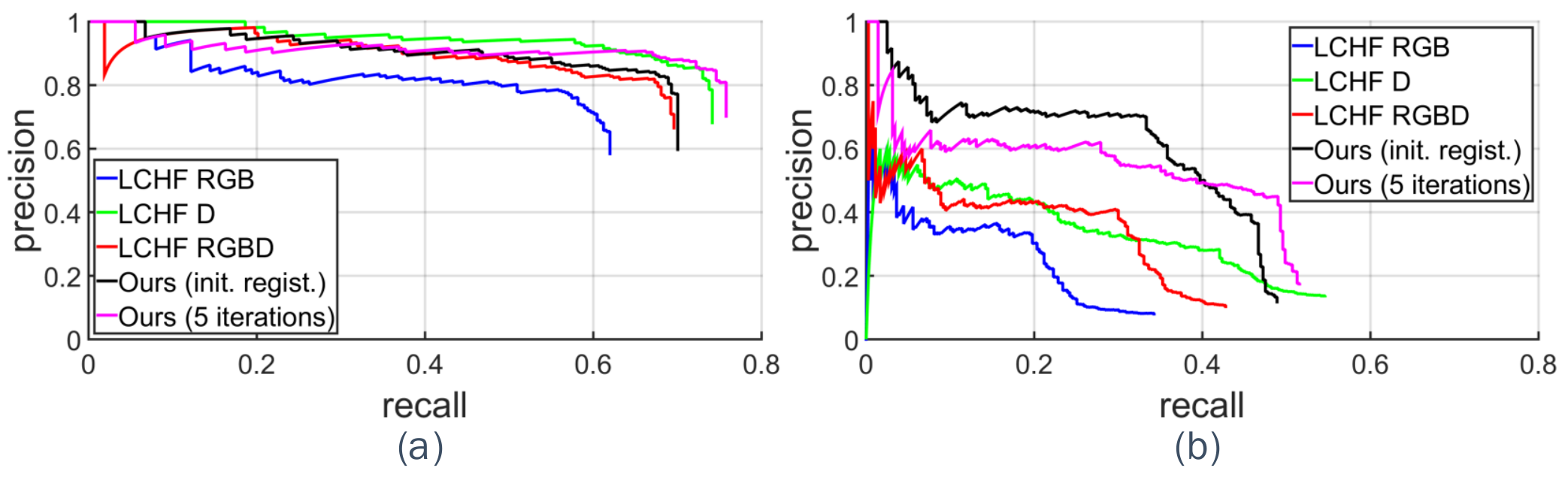}
\caption{PR curves of the \enquote{coffee cup} (left) and of the \enquote{camera} (right) dataset obtained from the interclass experiments: each image compares our method (initial registration and iterative pose refinement) with the LCHFs trained separately on the RGB, D and RGBD channels. F1 scores are presented in Table \ref{table_comparison}.}
\label{fig5}
\end{figure}
The last parameter we test in the intraclass experiments is the iteration number. We test several \textit{Iterative Hough Forests with Histogram of Control Points} each of which has $k = 0, 1, 3,$ and $5$ iterations, respectively. Their PR curves are shown in Fig. \ref{fig4} (c). As expected, the forests that use greater number of iterations show better performances (see Table \ref{table_params}) since more discriminative features are extracted thanks to the noise removal process.
\begin{figure*}[!t]
\captionsetup[subfigure]{labelformat=empty}
\centering
\includegraphics[height=2.2in]{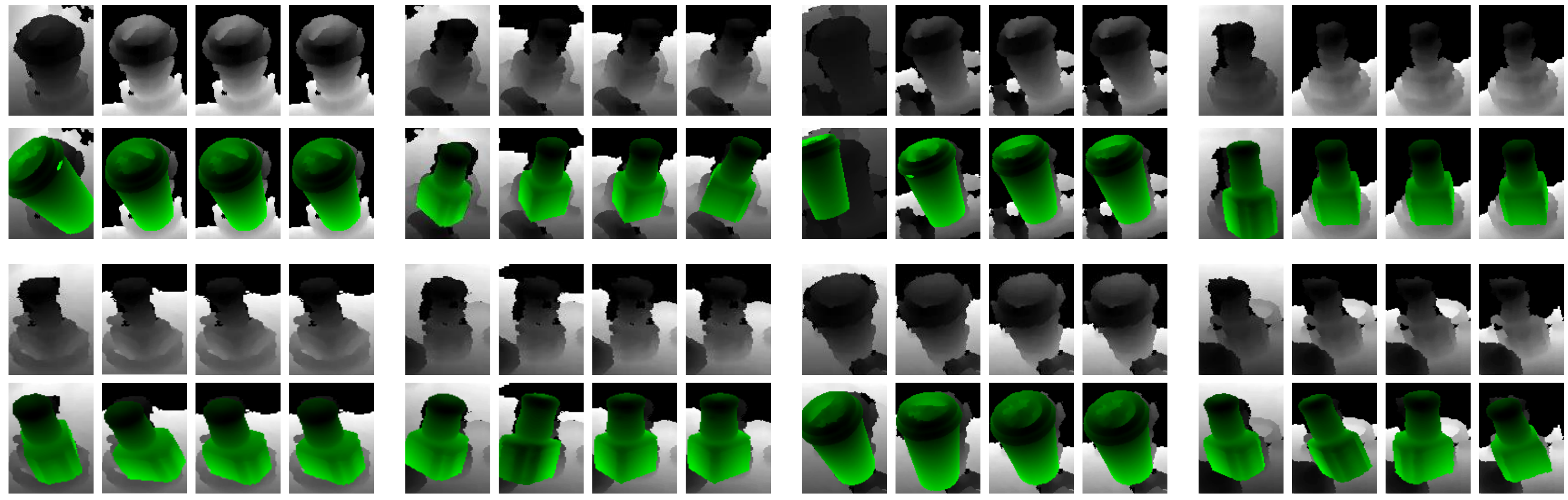}
\caption{Some qualitative results. For each octonary: the $1^{st}$ column illustrates the test image and the initial hypothesis (\textit{initial registration}), and the remaining columns demonstrate the $1^{st}$, the $3^{rd}$ and the $5^{th}$ iterations (\textit{iterative pose refinement}). The test images are updated by removing background/foreground clutter.}
\label{fig6}
\end{figure*}
\subsection{Interclass Experiments}
These experiments are conducted on the \enquote{coffee cup} and the \enquote{camera} datasets to compare our approach with the state-of-the-art methods including the Latent-Class Hough forests (LCHF) \cite{7} trained separately on the color gradient (LCHF-RGB), the surface normal (LCHF-Depth) and the color gradient + the surface normal (LCHF-RGBD) features. In order to make a fair comparison between methods, we train and test these versions of the LCHF by using the authors' software. The forest parameters are the same as our own approach.\\
\indent According to the F1 scores in Table \ref{table_comparison}, we observe that the LCHF trained on the color gradient features underperforms other methods. The main reason of this underperformance is the distortion along the object borders arising from the occlusion and the clutter, that is, the distortion of the color gradient information in the test process. When we train the LCHF by only using the depth information, we infer that the surface normals outperform the color gradients. The combined utilization of the color gradients and the surface normals in the LCHF produces approximately the same results as the LCHF-Depth. Our approach with the iterative pose refinement outperform other methods. Regarding the 'camera' object, we observe that the registration performances of all methods are relatively decreased. This is mainly because of that this object has large amount of missing depth pixels in addition to severe occlusion and clutter. Figure \ref{fig6} illustrates several qualitative results of our approach on the camera and the coffee cup objects \footnote{Supplementary video: http://www.iis.ee.ic.ac.uk/ComputerVision/}.
\begin{table}[!t]
\caption{F1 scores determined for different patch sizes, IBS resolution (N) \& feature dimension (d) and number of iteration}
\centering
\begin{tabular}[t]{|c|c||c|c|}
\hline
$\textbf{Patch}$&$ \textbf{F1}$   & $\textbf{N}$ \& $\textbf{d}$ &$ \textbf{F1}$\\
$\textbf{Size}$ &$ \textbf{Score}$& $ $                          &$ \textbf{Score}$\\
\hline
$\frac{1}{5}$         &0.5966           &80 \& 128                         &0.7068\\
\hline
$\frac{1}{4}$         &0.6096           &80 \& 256                         &0.7368\\
\hline
$\frac{1}{3}$         &0.6532           &80 \& 512                         &0.7425\\
\hline
$\mathbf{\frac{1}{2}}$&$\mathbf{0.7068}$&100 \& 128                        &0.6870\\
\hline
$\frac{2}{3}$         &0.6341           &$\mathbf{100}$ \& $\mathbf{256}$  &$\mathbf{0.7510}$\\
\hline
$\frac{3}{4}$         &0.6539           &100 \& 512                        &0.7438\\
\hline
\end{tabular}
\begin{tabular}[t]{|c|c|}
\hline
$\textbf{\#}$          &$ \textbf{F1}$    \\
$ \textbf{iter} $      &$ \textbf{Score}$ \\
\hline
0            &0.7510           \\
\hline
1            &0.7742           \\
\hline
3            &0.7745           \\
\hline
$\mathbf{5}$ &$\mathbf{0.7932}$\\
\hline
\end{tabular}
\label{table_params}
\end{table}
\begin{table}[!t]
\caption{F1 scores of the \enquote{coffee cup} and the \enquote{camera} datasets are shown. In both datasets our approach with the iterative pose refinement outperforms.}
\centering
\begin{tabular}[t]{|c|c|c|}
\hline
$\textbf{Approach}$ & $\textbf{Coffee Cup}$ & $\textbf{Camera}$         \\
\hline
LCHF-RGB                       &0.6595               & 0.2478           \\
\hline
LCHF-Depth                     &0.7860               & 0.3386           \\
\hline
LCHF-RGBD                      &0.7390               & 0.3456           \\
\hline
Ours (init. reg.)      &0.7510               & 0.4534           \\
\hline
Ours (iter. pose ref.) & $\mathbf{0.7932}$   & $\mathbf{0.4693}$\\
\hline
\end{tabular}
\label{table_comparison}
\end{table}
\section{Conclusion}
\label{conclusion}
In this study, we have proposed a novel architecture, \textit{Iterative Hough forest with Histogram of Control Points}, for 6 DoF object registration from depth images. We have introduced the Histogram of the Control Points, a scale-variant patch representation, and have encoded their rich discriminative information into the random forests. We train our forest using only the positive samples. During testing, we first roughly align the object and then iteratively refine this alignment. The experimental results report that our approach show better registration performance than the state-of-the-art methods. In the future, we plan to engineer a variable patch size approach and integrate it into the proposed iterative Hough forest architecture for further exploitation of the rich discriminative information provided by the HoCP features.

\end{document}